\documentclass[10pt]{article} 

\usepackage[accepted]{rlj} 

\usepackage{amssymb}            
\usepackage{graphicx}           
\usepackage{url}                
\usepackage{amsthm}             
\usepackage{bm}                 

\newtheorem{theorem}{Theorem}

\usepackage{listings}
\title{A-3PO: Accelerating Asynchronous LLM Training with Staleness-aware Proximal Policy Approximation}

\setrunningtitle{A-3PO: Accelerating Asynchronous LLM Training}

\author{Xiao-Can (Bruce) Li\textsuperscript{1,$\dagger$}, Shi-Liang (Bruce) Wu\textsuperscript{1,$\dagger$}, Zheng Shen\textsuperscript{1}}

\emails{hsiaotsan.li@alumni.utoronto.ca, \ \{okwsl201210, zhengshencn\}@gmail.com}

\affiliations{
$^{1}$\textbf{Huawei Canada}
\par
$^\dagger$ Equal contribution
}

\contribution{
    A staleness-aware proximal probability interpolation method that eliminates the computation cost of proximal policies in decoupled loss while retaining PPO's trust-region structure. We provide theoretical analysis establishing the sandwich property and contractive stability of the approximation.
    }
    {
    Decoupled loss \citep{hilton2022batch} introduced the proximal policy concept for asynchronous RL; our work approximates it rather than computing it explicitly.
    }

\contribution{
    Empirical evaluation across two model scales (1.5B and 8B parameters) demonstrates that our method achieves up to 1.8$\times$ speedup in training time while maintaining comparable task performance and superior training stability compared to both the standard decoupled GRPO and synchronous training baselines.
    }
    {
    Experiments are conducted on mathematical reasoning tasks (GSM8K and DAPO-Math-17k) using the large-scale RL framework AReaL \citep{fu2025areal}.
    }

\contribution{
    Open-source implementation contributed to the AReaL framework, providing an efficient large-scale asynchronous RL-based LLM post training algorithm.
    }
    {
    The implementation builds on the existing AReaL framework \citep{fu2025areal} and requires only minimal code changes (element-wise tensor operations) to the training loop.
    }


\summary{Decoupled PPO has been a successful reinforcement learning (RL) algorithm to deal with the high data staleness under the asynchronous RL setting. Decoupled loss used in decoupled PPO improves coupled-loss style of algorithms' (e.g., standard PPO, GRPO) learning stability by introducing a proximal policy to decouple the off-policy correction (importance weight) from the policy update constraint (trust region). However, the proximal policy requires an extra forward pass through the model at each training step, creating a computational overhead for large language models training. We observe that since the proximal policy only serves as a trust region anchor between the behavior and target policies, we can approximate it through simple interpolation without explicit computation. We call this approach A-3PO (APproximated Proximal Policy Optimization). A-3PO eliminates this overhead, accelerating training by 1.8$\times$ speedup while maintaining comparable performance. Code \& off-the-shelf example are contributed to the open-source RL training system AReaL: \url{https://github.com/areal-project/AReaL/tree/v1.0.0.rc1}.}

\begin{document}

\makeCover  
\maketitle  

\begin{abstract}
Decoupled PPO has been a successful reinforcement learning (RL) algorithm to deal with the high data staleness under the asynchronous RL setting. Decoupled loss used in decoupled PPO improves coupled-loss style of algorithms' (e.g., standard PPO, GRPO) learning stability by introducing a proximal policy to decouple the off-policy correction (importance weight) from the policy update constraint (trust region). However, the proximal policy requires an extra forward pass through the model at each training step, creating a computational overhead for large language models training. We observe that since the proximal policy only serves as a trust region anchor between the behavior and target policies, we can approximate it through simple interpolation without explicit computation. We call this approach A-3PO (APproximated Proximal Policy Optimization). A-3PO eliminates this overhead, accelerating training by 1.8$\times$ speedup while maintaining comparable performance. Code \& off-the-shelf example are contributed to the open-source RL training system AReaL: \url{https://github.com/areal-project/AReaL/tree/v1.0.0.rc1}.
\end{abstract}

\section{Introduction}
Reinforcement learning (RL) \citep{sutton1998reinforcement} has become a central approach to improve the reasoning capabilities of large language models (LLMs) \citep{grpo-deepseek-math,yang2025qwen3technicalreport,zheng2025gspo,yu2025dapoopensourcellmreinforcement,ouyang2022instructgpt}, with extensive surveys covering RL from human feedback methods and workflows \citep{kaufmann2025rlhfsurvey,dong2024rlhfworkflow,wang2024secrets} and various alternative approaches including AI feedback \citep{lee2024rlaif} and safety considerations \citep{dai2024safe}. Among RL algorithms, Proximal Policy Optimization (PPO) \citep{schulman2017proximal} has emerged as the dominant method due to its stable trust-region constraints, building upon earlier trust region methods like TRPO \citep{schulman2015trust}. However, standard PPO performs rollout-then-training loop, i.e., the training stage must wait until the rollout stage collects predefined number of episodes, limiting throughput (measured by the number of environment steps per unit of time) and under-utilizes computational resources.

To improve the throughput and computational resources utilization, asynchronous RL \citep{fu2025areal,roll_wang2025reinforcement,verl_sheng2024hybridflow,feng2025mindspeedrldistributeddataflow,noukhovitch2024asyncrlhf,wu2025llamarl,espeholt2018impala,lan2025asyncfed,duan2024distributed,cohen2025soft} treats rollout and training as two independent engines, which can be executed in parallel. Nevertheless, the target policy on the training engine can be several updates ahead of the behavior policy on the rollout engine. Such staleness (off-policyness) caused by asynchronous RL setting could lead to severe learning instability in standard PPO. To mitigate this, decoupled PPO \citep{hilton2022batch} improves the learning stability by introducing a proximal policy that decouples the off-policy correction (importance weight) from the policy update constraint (trust region). Decoupled loss empirically demonstrates improved learning stability in Atari games when high off-policyness exists. Apart from Atari games, AReaL \citep{fu2025areal}, an LLM post training framework, demonstrated superior learning stability of decoupled loss on LLM reasoning tasks under high off-policyness setting. Thanks to asynchronous RL setup, AReaL also achieved up to 2.77$\times$ training speedup. However, the proximal policy in decoupled loss requires an extra forward pass through the neural network at each training step, which is expensive for autoregressive LLMs. This overhead limits the potential speedups from asynchronous training.

\textbf{This raises a natural question: do we really need to compute the proximal policy explicitly? Looking at the objective from first principles, the proximal policy simply serves as a trust region anchor: it does not need to be computed from the network, but it just needs to lie somewhere between the behavior and target policies to prevent extreme importance weights.} This insight leads to our solution: instead of computing the proximal policy through a forward pass, we approximate it by interpolating between the behavior policy and the target policy in log-probability space. Our staleness-aware interpolation weighs fresher data more heavily, maintaining the stabilizing effect of decoupled loss while eliminating the computational overhead.

Our contributions are threefold:
\begin{enumerate}
    \item A staleness-aware proximal probability interpolation method that eliminates the computation cost of proximal policies in decoupled loss while retaining PPO's trust-region structure.
    \item Empirical evaluation across two model scales (1.5B and 8B parameters) demonstrates that our method achieves up to 1.8$\times$ speedup in training time while maintaining comparable task performance and superior training stability compared to both the standard decoupled GRPO and synchronous GRPO training baselines, with particular advantages at larger scales.
    \item Open-source implementation, providing an efficient large-scale asynchronous RL-based LLM post training algorithm.
\end{enumerate}

\section{Preliminary}

\subsection{From Coupled Loss to Decoupled Loss}

Among coupled-loss RL algorithms, the standard Proximal Policy Optimization (PPO) \citep{schulman2017proximal} has become one of the most dominant methods, thanks to its stable trust-region constraints that prevent destructive policy updates. In standard PPO, the clipped objective is:
\begin{align}
L^{\text{CLIP}}_{\text{coupled}}(\theta) = \mathbb{E}_t \left[ \min \left( r_t(\theta) \hat{A}_t, \text{clip}(r_t(\theta), 1-\epsilon, 1+\epsilon) \hat{A}_t \right) \right], \label{eq:coupled-loss}
\end{align}
where $r_t(\theta) = \frac{\pi_\theta(a_t|s_t)}{\pi_{\text{old}}(a_t|s_t)}$ is the importance weight and $\hat{A}_t$ is the advantage estimate. Here, the old policy $\pi_{\text{old}}$ serves two purposes simultaneously: it provides importance sampling weights \citep{papini2024active,deasis2023valueaware} to correct for off-policy data, and it defines the trust region that constrains how far the new policy can deviate.

However, this coupled role becomes problematic in asynchronous RL settings, where the behavior policy used for data collection may be several updates behind the current policy. When training on stale data, using the behavior policy as the trust region anchor pulls the current policy towards outdated, potentially lower-quality policies, leading to learning instability.

Decoupled loss \citep{hilton2022batch} addresses this issue with a key insight: these two roles of $\pi_{\text{old}}$ can and should be separated. For importance sampling, we must use the actual behavior policy $\pi_{\text{behav}}$ that generated the data. But for trust region control, we can use some recent policy $\pi_{\text{prox}}$ as the proximal anchor. The decoupled clipped objective becomes:

\begin{align}
L^{\text{CLIP}}_{\text{decoupled}}(\theta) = \mathbb{E}_t \Big[
 \underbrace{ \frac{\pi_{\text{prox}}(a_t|s_t)}{\pi_{\text{behav}}(a_t|s_t)} }_{\text{Importance Weight}}
 \min\!\bigg(
  \frac{\pi_\theta(a_t|s_t)}{\pi_{\text{prox}}(a_t|s_t)} \hat{A}_t,\,
  \text{clip}\!\Big(
   \underbrace{ \frac{\pi_\theta(a_t|s_t)}{\pi_{\text{prox}}(a_t|s_t)} }_{\text{Trust Region Anchor}},\,
   1-\epsilon,\,
   1+\epsilon
  \Big) \hat{A}_t
 \bigg)
\Big] \label{eq:decoupled-loss}
\end{align}

By using a more recent policy $\pi_{\text{prox}}$ as trust region anchor, the trust region now constrains updates around a higher-quality policy, while still correctly accounting for the off-policy nature of the data through $\pi_{\text{behav}}$. This decoupling is crucial for asynchronous RL systems like AReaL \citep{fu2025areal}, which achieves significant training speedup by completely separating generation and training phases.

\subsection{The Price of Decoupled Loss}

While decoupled loss improves stability, it comes with a computational price. At the start of each training step, we must perform a forward pass through the model to compute $\pi_{\text{prox}}$. This proximal policy is then frozen (detached from gradient updates) and used as the trust region anchor for all subsequent mini-batch updates within that training step. This overhead becomes significant for large language models where a single forward pass can take 10 seconds or longer, depending on the input size, model size, and hardware specification. The timing is demonstrated in Figure~\ref{fig:recompute_time}.

\section{Methodology: A-3PO}

In our proposed method, the objective function is still Eq. \ref{eq:decoupled-loss}, but the proximal policy is interpolated between the behavior policy and target policy at log scale \citep{neal1998annealedimportancesampling}, weighted by a staleness-aware coefficient \citep{asyncSGD_tan2024}, as defined in Eq. \ref{eq:loglinear-approximation}. The more the target policy is ahead of the behavior policy (high staleness), the closer the approximated proximal policy is to the target policy:
\begin{align}
\log \pi_{\text{prox}} = \alpha\log \pi_{\text{behav}} + (1-\alpha) \log \pi_\theta, \label{eq:loglinear-approximation}
\end{align}

Input arguments $(a|s)$ are omitted for brevity. The main reason to perform interpolation in log-probability space rather than probability space is that it maintains numerical stability: log-probabilities avoid underflow issues that arise when dealing with very small probability values common in large action spaces. Practically, the inference engine, e.g., SGLang \citep{zheng2024sglangefficientexecutionstructured} and vLLM \citep{vllm_kwon2023efficient}, provides token log-probabilities by default.

Here, $\alpha$ is a staleness-aware coefficient depending on the staleness $d$ (the training step difference between target and behavior policies). $v(\pi)$ denotes the training step of policy $\pi$:
\begin{align}
    d &= v(\pi_{\theta}) - v(\pi_{\text{behav}}), \quad
\alpha =
\begin{cases}
1, & d =0, \\[3pt]
\dfrac{1}{d}, & d \geq 1.
\end{cases}
\end{align}

When $d=0$, it recovers the standard synchronous PPO, hence the approximated proximal policy is exactly the behavior policy.

When $d \geq 1$, the coefficient $\alpha$ monotonically decreases as the staleness $d$ increases, approximating the proximal policy more with the target policy, and give less weights to the behavior policy.

\subsection{Theoretical Stability Analysis}

\subsubsection{Sandwich Property} Thanks to the interpolation, $\pi_{\mathrm{prox}}(a | s)$ is bounded:
    \begin{align}
    \min\{\pi_{\mathrm{behav}}(a | s), \pi_\theta(a | s)\}
    \le
    \pi_{\mathrm{prox}}(a | s)
    \le
    \max\{\pi_{\mathrm{behav}}(a | s), \pi_\theta(a | s)\}.
    \end{align}
This property guarantees that $\pi_{\mathrm{prox}}$ serves as a valid trust-region anchor under policy staleness.

\subsubsection{Contractive Stability}
The staleness-aware proximal policy defined by log-linear interpolation admits a simple closed-form importance weight:
\begin{align}
r(a | s) \overset{\Delta}{=}
\frac{\pi_\theta(a | s)}{\pi_{\mathrm{prox}}(a | s)} = \Big(\dfrac{\pi_{\theta}(a | s)}{\pi_{\text{behav}}(a | s)}\Big)^{\alpha}
\end{align}

This form implies that importance weights are contractively scaled as staleness increases, preventing extreme ratios and ensuring stable trust-region updates. Moreover, raising importance weights to a power $\alpha < 1$ provably vanishes their variance, leading to more stable and fewer clipped updates in PPO-style objectives. A unified theoretical analysis establishing bounded updates and contractive stability is provided in Appendix~\ref{app:theoretical}.

\subsection{Implementation}

The core implementation of our staleness-aware proximal policy approximation is remarkably simple. In Listing~\ref{lst:implementation} we show the key computation in PyTorch, extracted from our contribution to the AReaL framework \citep{fu2025areal}.

\begin{lstlisting}[caption={Core implementation of staleness-aware proximal policy approximation.}, label=lst:implementation, float=htbp]
def compute_prox_logp_approximation(
    old_logp: torch.Tensor,      # (*@$\log \pi_{\text{behav}}$@*)
    logprobs: torch.Tensor,      # (*@$\log \pi_\theta$@*)
    versions: torch.Tensor,      # (*@$v(\pi_{\text{behav}})$@*) per token
    current_version: int,        # (*@$v(\pi_\theta)$@*)
) -> torch.Tensor:
    """Approximate proximal policy log-probabilities."""
    # Extract version information
    v_behave = versions.float()
    v_theta = float(current_version)

    # Compute staleness: (*@$d = v(\pi_\theta) - v(\pi_{\text{behav}})$@*)
    (*@\colorbox{yellow!30}{staleness = v\_theta - v\_behave}@*)

    # Compute staleness-aware coefficient (*@$\alpha$@*)
    # (*@$\alpha = 1$@*) when (*@$d =0$@*), (*@$\alpha = 1/d$@*) when (*@$d \geq 1$@*)
    (*@\colorbox{yellow!30}{alpha =}@*) torch.where(
        staleness >= 1,
        (*@\colorbox{yellow!30}{1.0 / staleness}@*),
        torch.ones_like(v_behave),
    )

    # Log-linear interpolation: (*@$\log \pi_{\text{prox}} = \alpha\log \pi_{\text{behav}} + (1-\alpha) \log \pi_\theta$@*)
    (*@\colorbox{yellow!30}{prox\_logp = alpha * old\_logp + (1 - alpha) * logprobs}@*)

    return prox_logp
\end{lstlisting}
The implementation highlights the efficiency of our approach: computing $\pi_{\text{prox}}$ requires only element-wise arithmetic operations on tensors that are already available from the training loop. No additional forward pass through the neural network is needed, making this approximation essentially free compared to the 10-second forward pass required for explicit computation.

\section{Experiments}
\subsection{Training details}

In this work, we focus on mathematical reasoning tasks to evaluate our algorithm, which can be
readily applied to other tasks. We adopt the AReaL framework \citep{fu2025areal} for training. We use GRPO with decoupled loss \citep{hilton2022batch} and synchronous GRPO \citep{grpo-deepseek-math} as our baseline algorithms and estimate advantages using group reward normalization.

We conduct experiments with two different model-dataset configurations:

\textbf{Setup 1: Qwen2.5-1.5B-Instruct on GSM8K.} We train and evaluate on GSM8K \citep{cobbe2021gsm8k}, a dataset of 8.5K grade school math problems that require 2-8 steps of multi-step reasoning using basic arithmetic operations. For rollout, the prompt batch size is 256 and we sample 4 responses for each prompt. We set the maximum response length as 1,024 tokens, and max tokens per mini batch is 10,240.

\textbf{Setup 2: Qwen3-8B on DAPO-Math-17k.} We also use the base model Qwen3-8B \citep{yang2025qwen3technicalreport} trained and evaluated on the DAPO-Math-17k dataset \citep{yu2025dapoopensourcellmreinforcement}. For rollout, the prompt batch size is 128 and we sample 4 responses for each prompt. We set the maximum response length as 2,048 tokens, and max tokens per mini batch is 10,240.

\textbf{Common hyperparameters.} For both setups, we use the Adam optimizer \citep{kinga2015method} with a constant learning rate of $8.5\times10^{-6}$. For training, the number of mini batches is set to 4, i.e., 4 gradient updates for each training step. For sampling parameters, the temperature is set to 1.0, and top-$p$ is set 1, and top-$k$ is set to use all vocabularies.

\subsection{Results \& Discussion}
We evaluate our A-3PO method (labeled as ``loglinear'') against two baselines: the asynchronous decoupled GRPO with proximal policy recomputing \citep{hilton2022batch} (labeled as ``recompute'') and synchronous GRPO (labeled as ``sync''). The synchronous baseline represents the standard coupled-loss approach without asynchronous training, serving as a reference for comparing both decoupled methods. We present results across both experimental setups that demonstrate consistent improvements in computational efficiency and training stability.

\subsubsection{Computational Efficiency}

\begin{figure}[htb]
\centering
\includegraphics[width=0.49\textwidth]{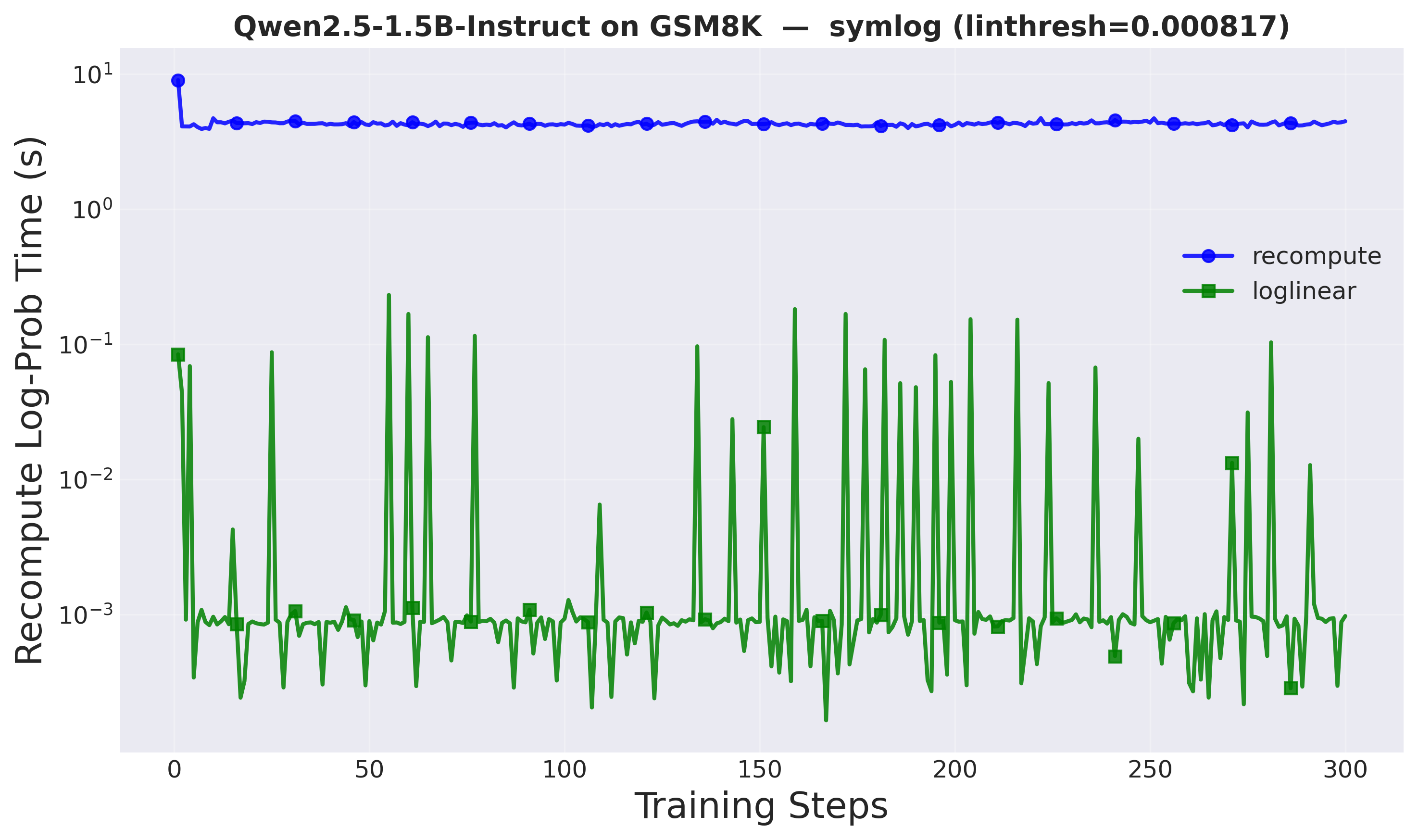}
\includegraphics[width=0.49\textwidth]{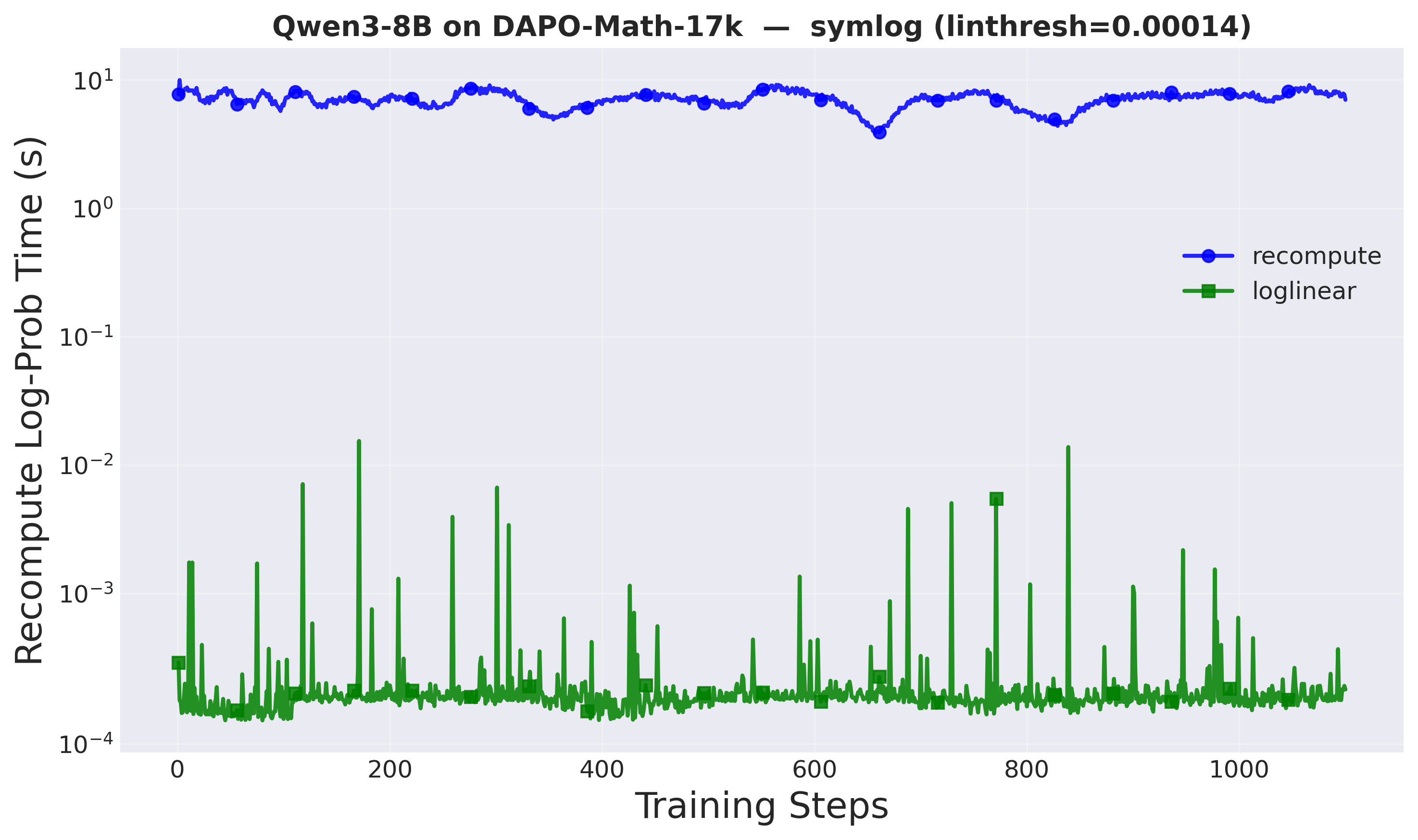}
\caption{Log probability computation time comparison. The loglinear method achieves near-instantaneous computation compared to the 10-second forward pass required by recomputing. Sync does not require this computation.}
\label{fig:recompute_time}
\end{figure}

Our staleness-aware approximation method dramatically reduces the computational cost of proximal policy evaluation. Figure~\ref{fig:recompute_time} shows the wall-clock time required to compute log probabilities of the proximal policy at each training step. The loglinear approximation method achieves near-zero computation time (mean: 0.0012 seconds), while the full recompute approach requires approximately 4 to 8 seconds per step depending on the setting, and the sync method does not require proximal policy computation. This represents at least 3,000$\times$ speedup in proximal policy's log probability computation for decoupled methods, translating directly to reduced training time.

\begin{figure}[htb]
\centering
\includegraphics[width=0.49\textwidth]{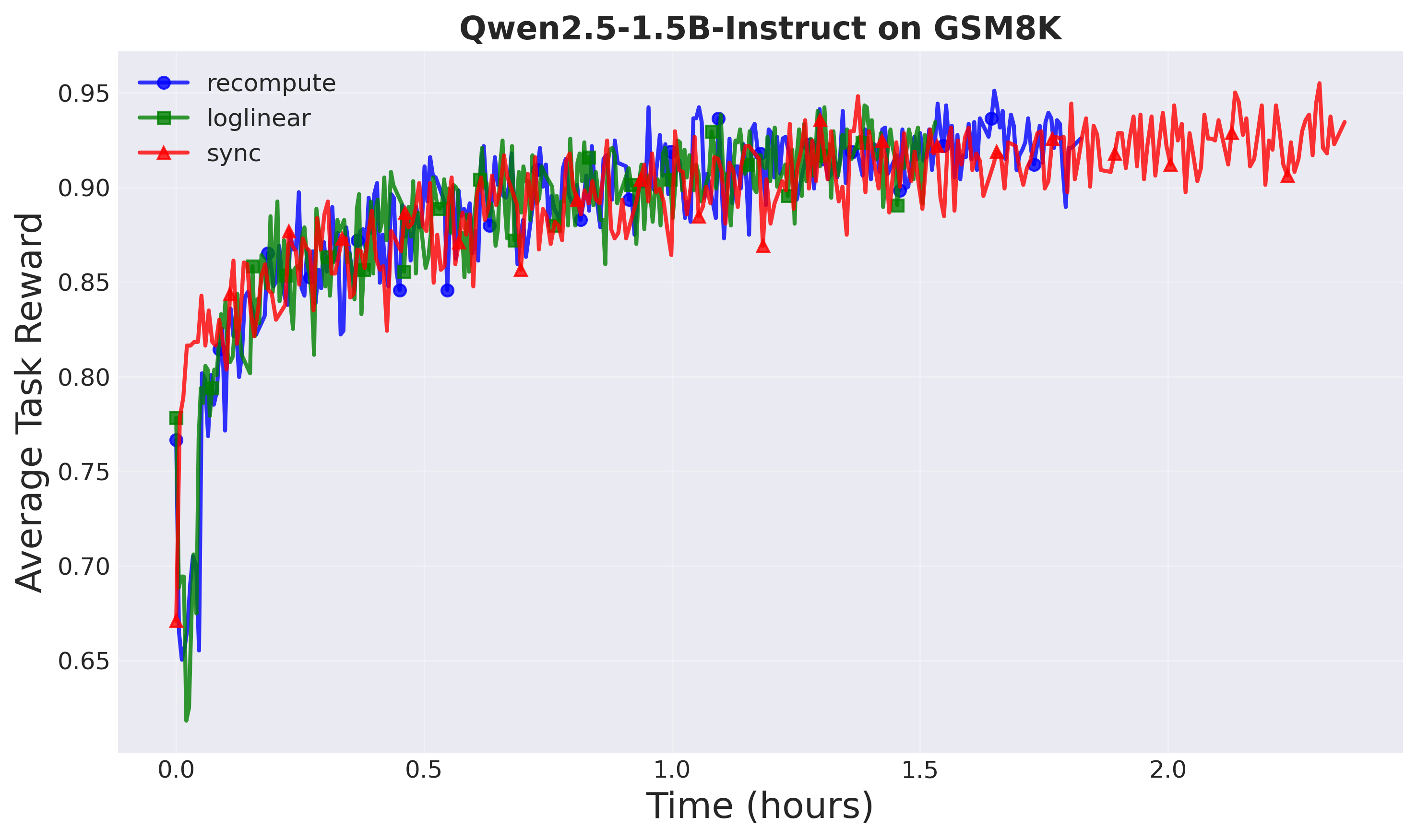}
\includegraphics[width=0.49\textwidth]{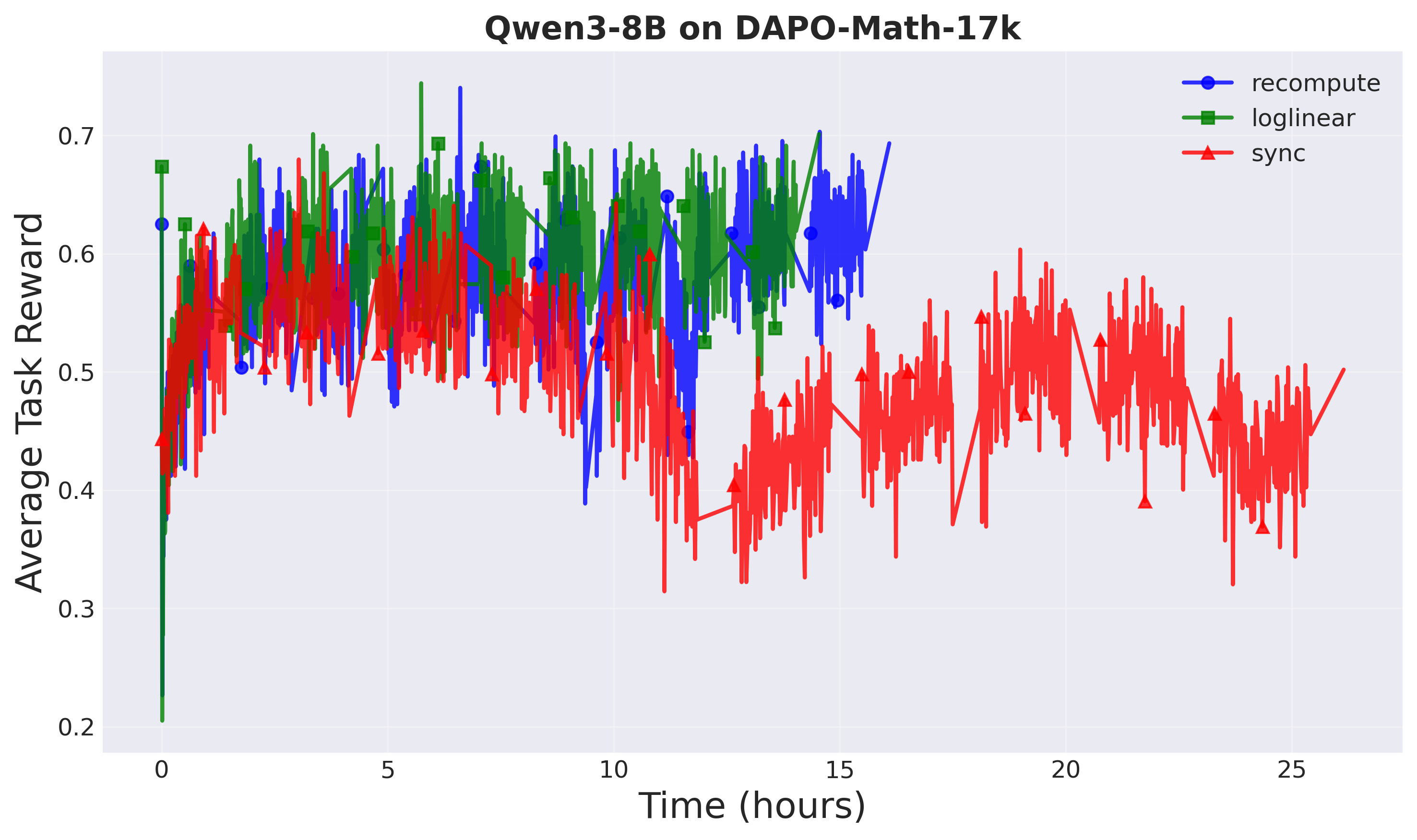}
\caption{Training progress (average task reward vs. wall-clock time). All curves use the same training epochs. Asynchronous training with loglinear approximation is fastest.}
\label{fig:reward_vs_time}
\end{figure}

As shown in Figure~\ref{fig:reward_vs_time}, the loglinear method consistently achieves faster training while maintaining comparable task performance to both baselines. For a fair comparison, the same training epochs are used for all methods. For both setups, our A-3PO method converges to similar final task rewards, demonstrating that our approximation maintains task performance while significantly reducing the training time across different model scales and datasets.

\begin{figure}[htb]
\centering
\includegraphics[width=0.49\textwidth]{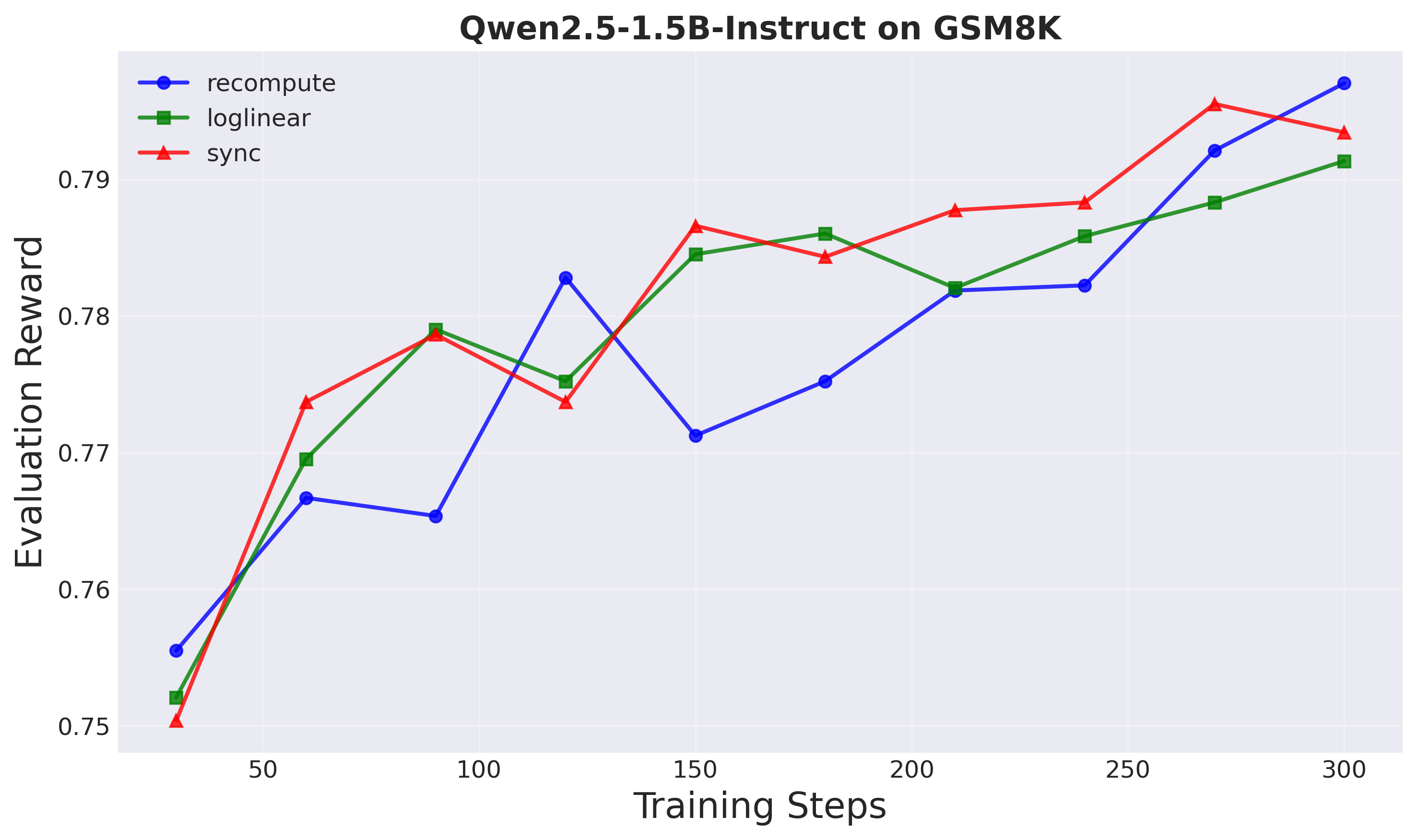}
\includegraphics[width=0.49\textwidth]{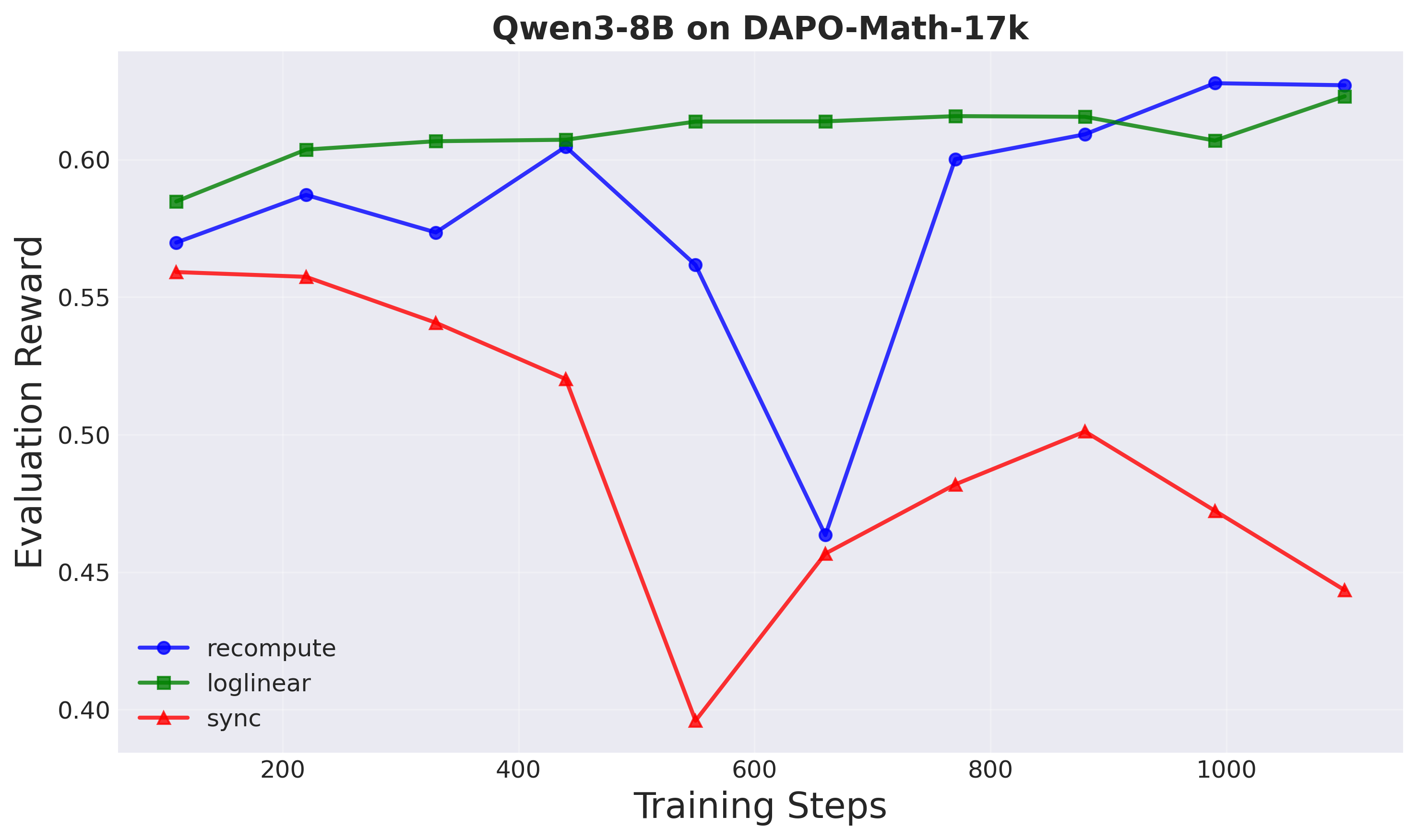}
\caption{Evaluation reward on held-out test prompts. Setup 1: all methods converge similarly. Setup 2: asynchronous methods substantially outperform sync, demonstrating decoupled loss effectiveness at larger scales.}
\label{fig:eval_rollout_reward}
\end{figure}

To further validate the training progress and model quality, we evaluate the policies on held-out test prompts during training. Figure~\ref{fig:eval_rollout_reward} shows the evaluation reward trajectories over training steps. In Setup 1, all three methods achieve comparable final evaluation rewards (gap $<$ 1\%), demonstrating that the loglinear approximation maintains task performance. In Setup 2, the asynchronous methods (loglinear and recompute) significantly outperform the baseline ``sync'', converging to evaluation rewards around 0.63 compared to 0.44 for sync method. This substantial gap highlights the benefits of asynchronous training with decoupled loss at larger model scales. Notably, the loglinear method achieves this performance without the computational overhead of explicit proximal policy computation, making it the most efficient approach overall.

\subsubsection{Training Stability Analysis}

\begin{figure}[htb]
\centering
\includegraphics[width=0.49\textwidth]{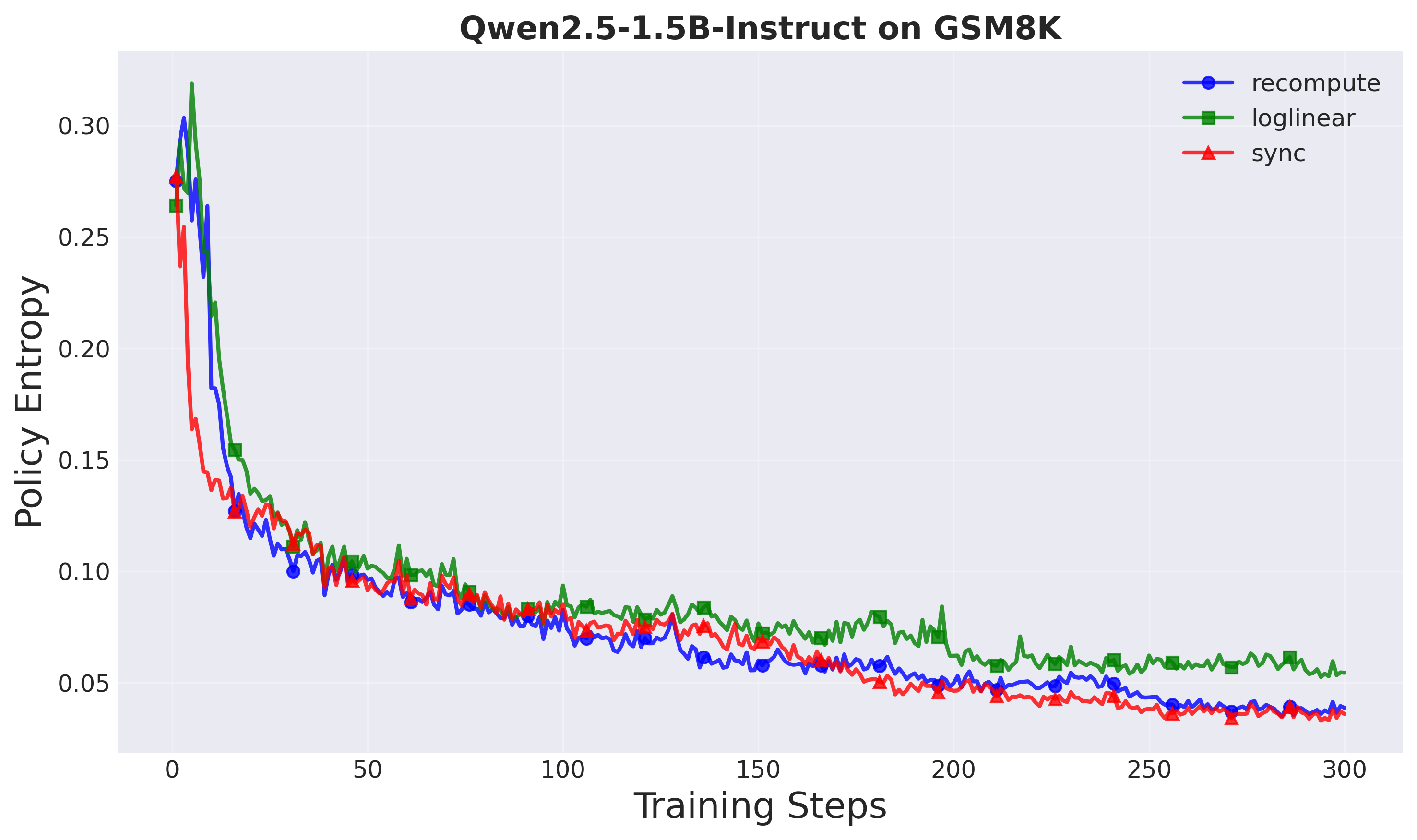}
\includegraphics[width=0.49\textwidth]{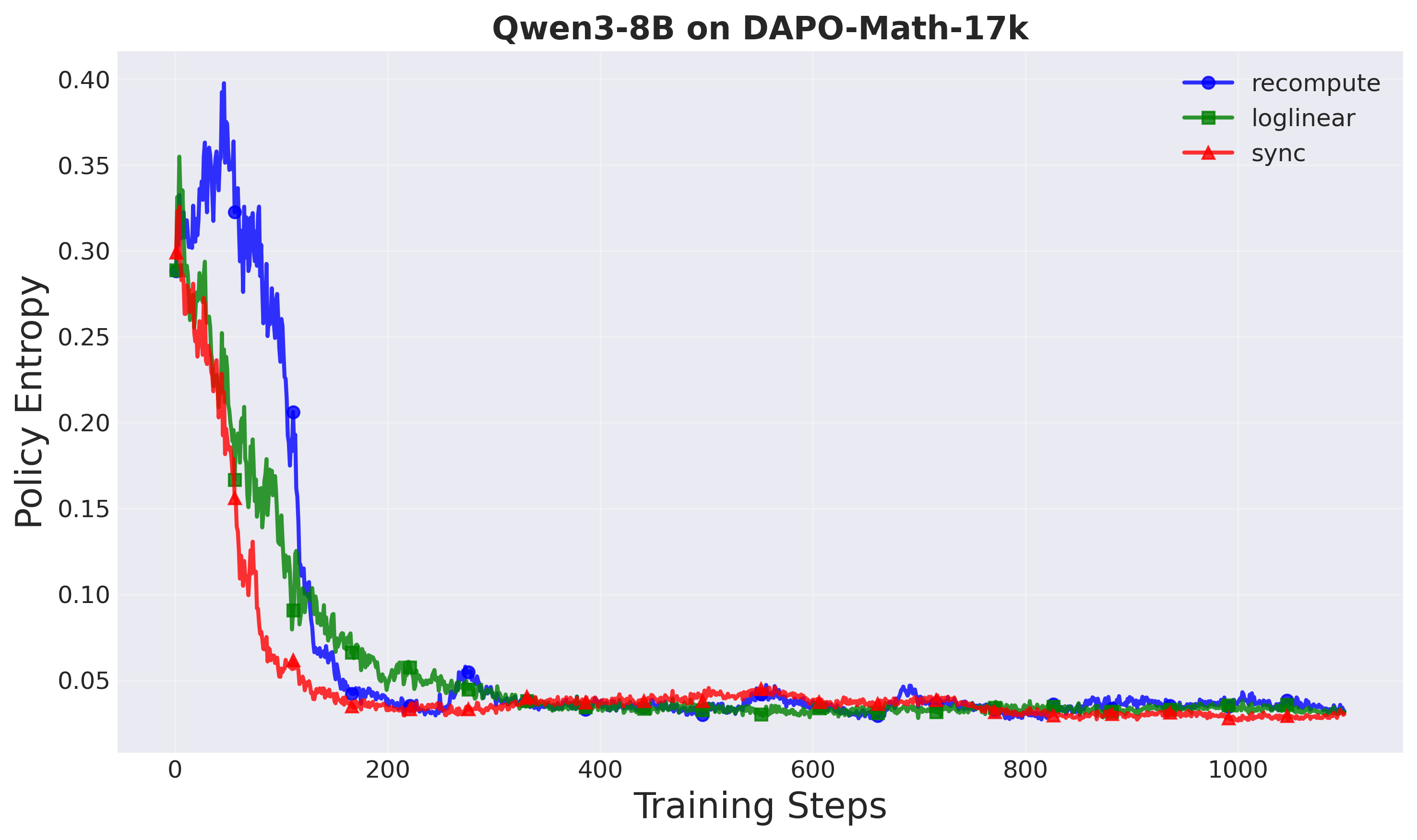}
\caption{Policy entropy over training steps. All methods show healthy entropy decay, indicating stable policy optimization.}
\label{fig:entropy}
\end{figure}

Beyond computational efficiency, we analyze whether our A-3PO maintains the stabilizing properties of decoupled loss across both experimental setups. Figure~\ref{fig:entropy} shows that all three methods exhibit similar entropy decay patterns. The baseline ``sync'' and both decoupled methods show healthy entropy decay, indicating stable policy optimization. This demonstrates that our approximation preserves healthy exploration dynamics across different model scales and datasets, comparable to both the recompute method and the synchronous baseline.

\begin{figure}[htb]
\centering
\includegraphics[width=0.49\textwidth]{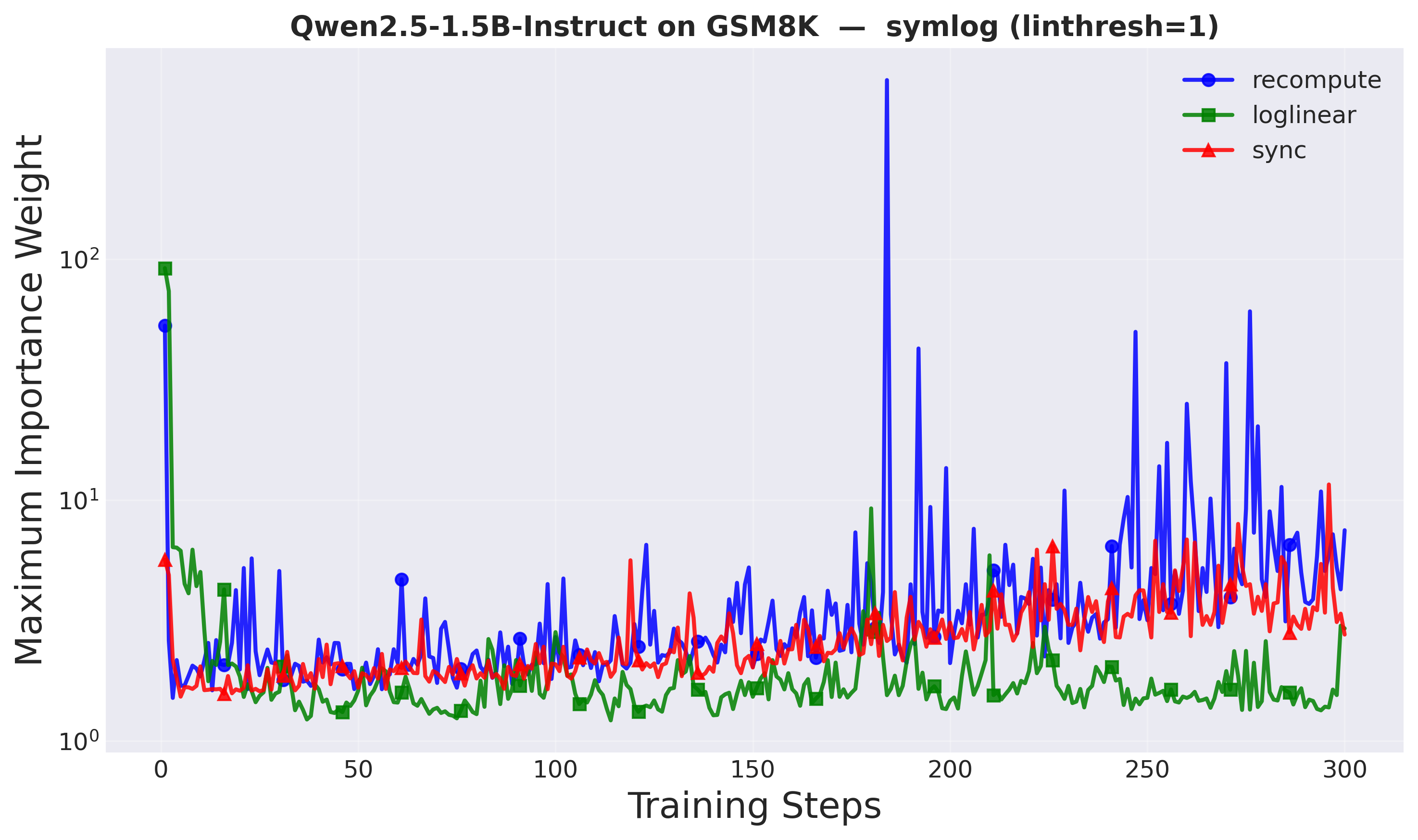}
\includegraphics[width=0.49\textwidth]{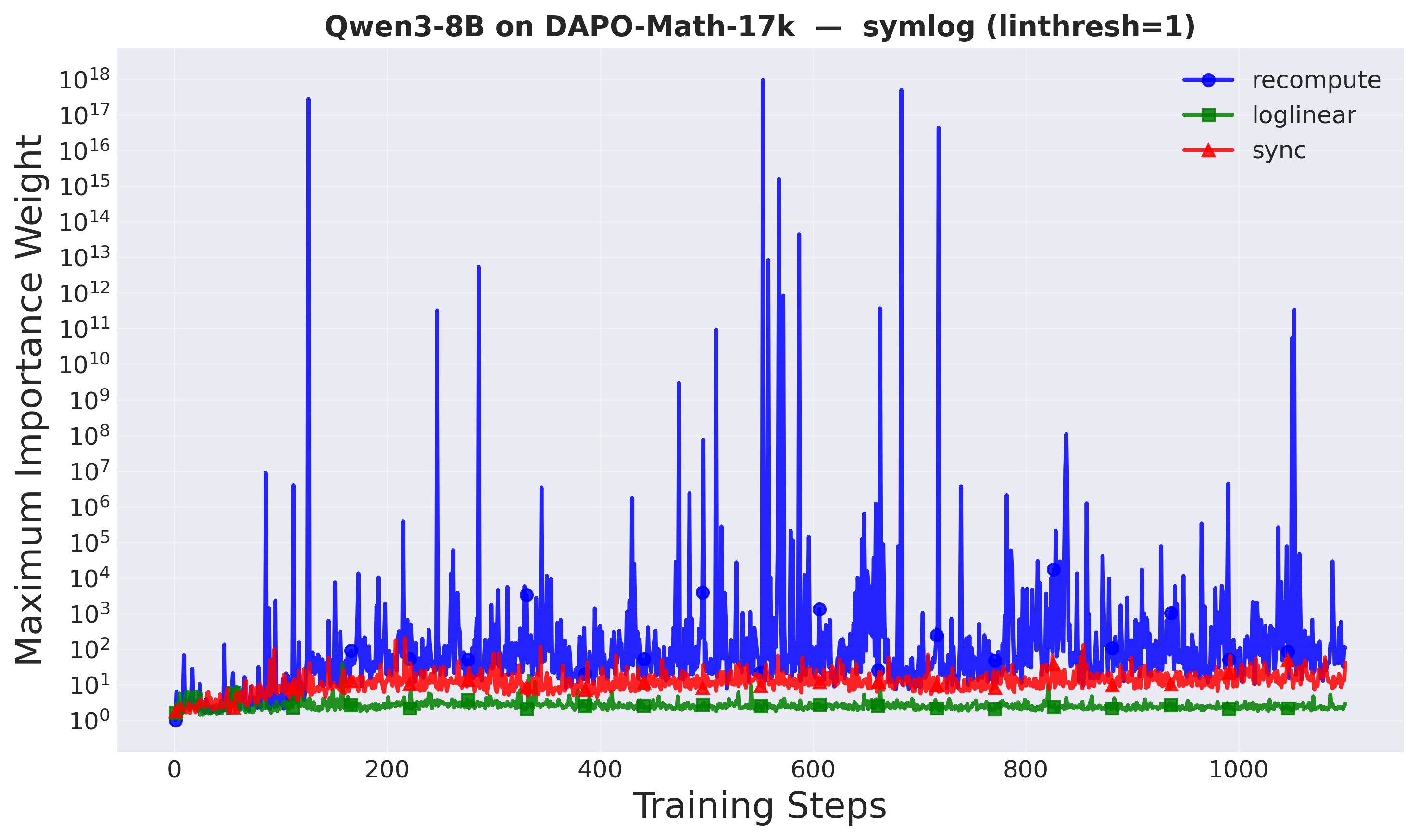}
\includegraphics[width=0.49\textwidth]{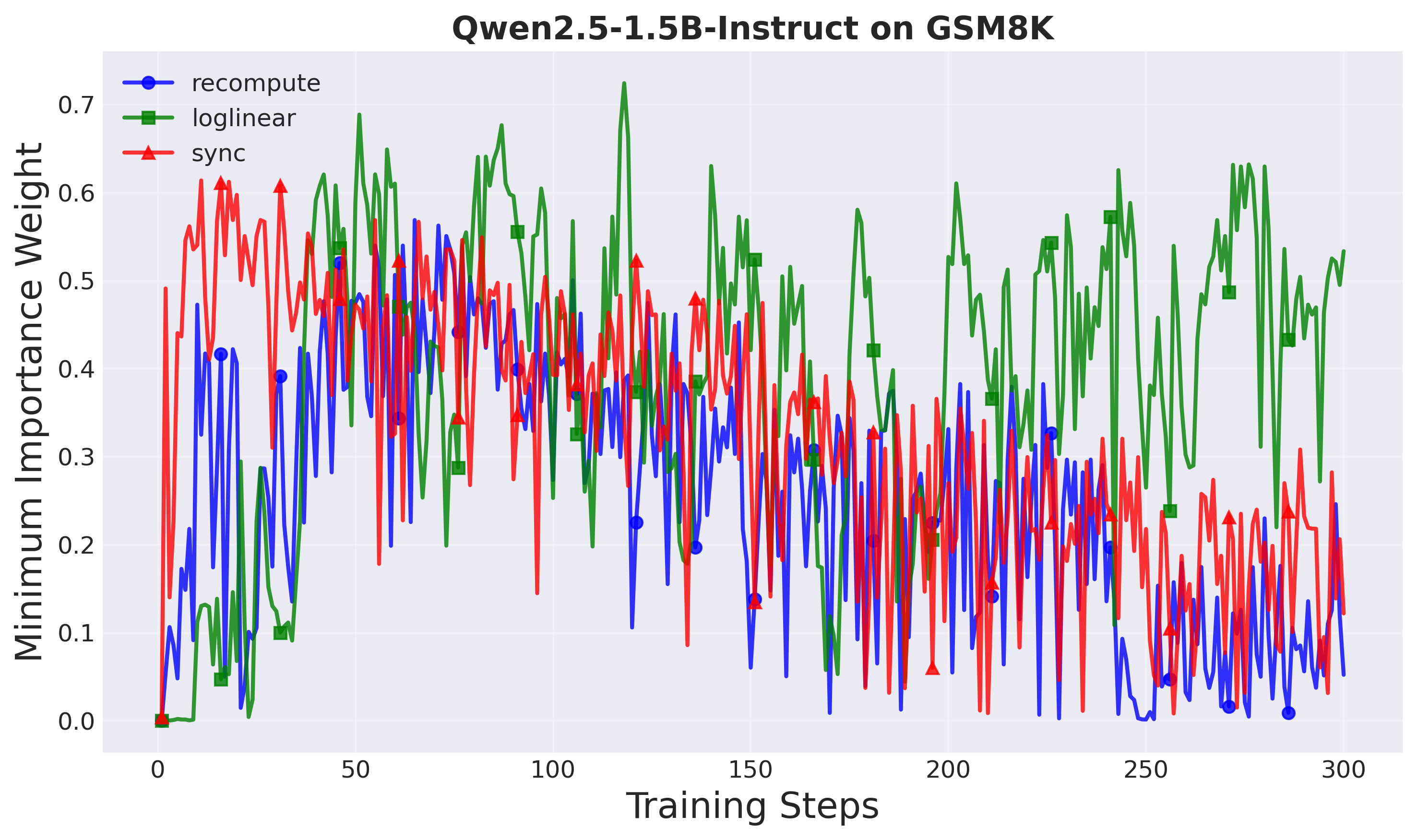}
\includegraphics[width=0.49\textwidth]{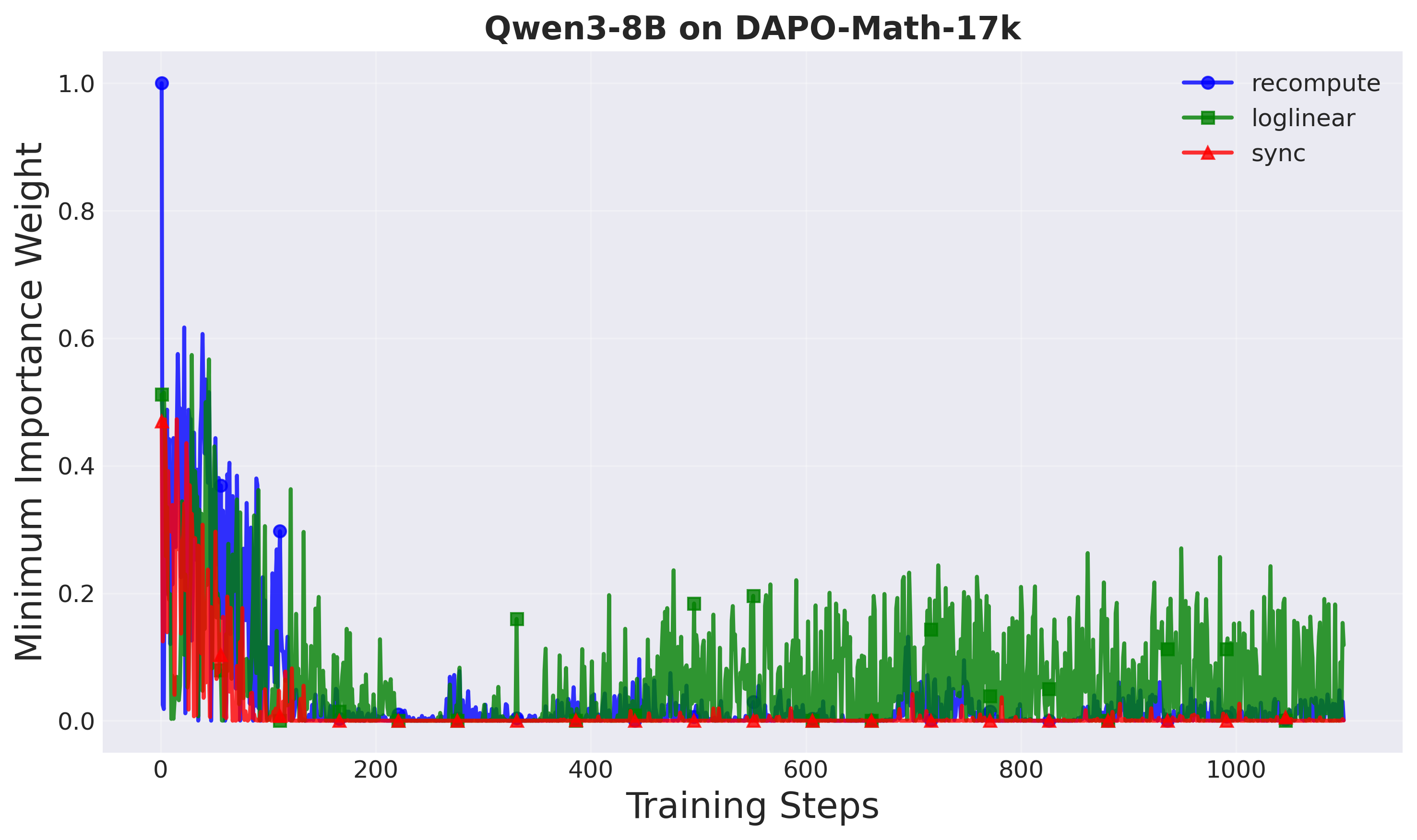}
\caption{Importance weight statistics (top: max; bottom: min). Loglinear shows more controlled weights. Setup 2: recompute produces very high weights, indicating instability at larger scales.}
\label{fig:importance_weights}
\end{figure}

Importance weight statistics provide further evidence of stability across both setups. Figure~\ref{fig:importance_weights} displays the maximum and minimum importance weights during training for the two decoupled methods (note that the sync method uses coupled loss and does not compute separate importance weights). Notably, the recompute approach exhibits very high importance weights, indicating that the recomputed proximal policy becomes unreliable at larger model scales. In contrast, the loglinear approximation consistently maintains more balanced importance sampling that avoids extreme weight values which can destabilize training.

\begin{figure}[htb]
\centering
\includegraphics[width=0.49\textwidth]{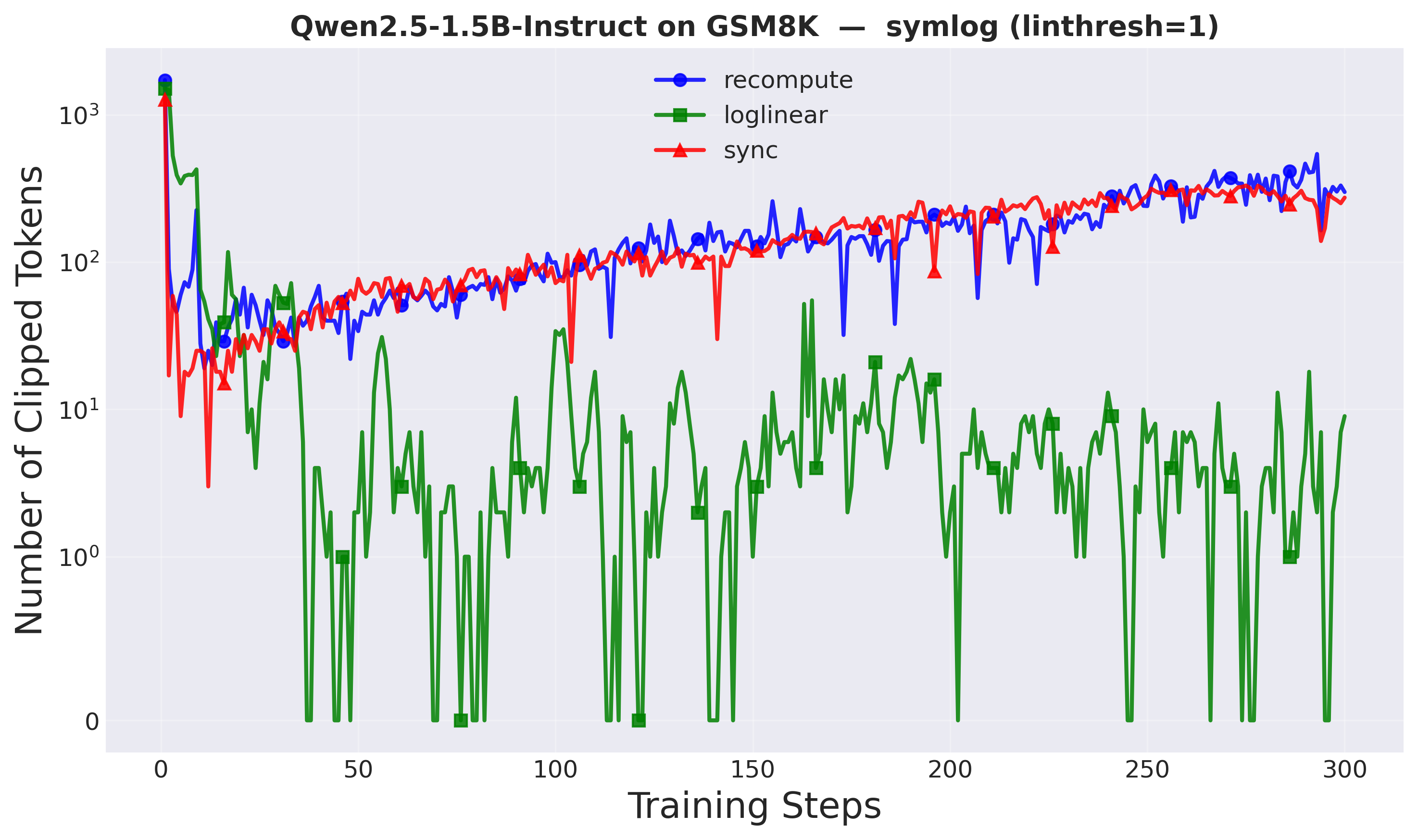}
\includegraphics[width=0.49\textwidth]{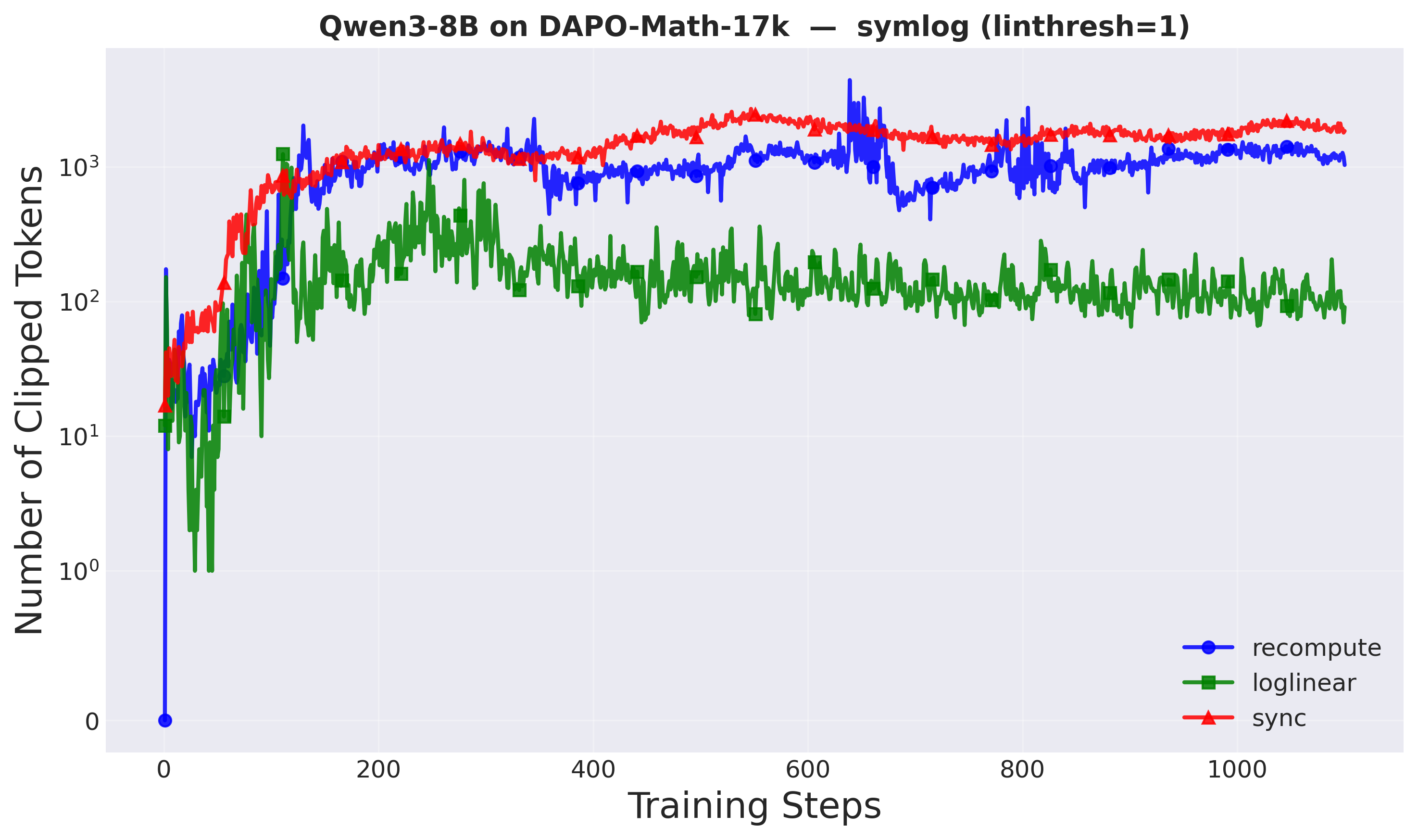}
\caption{Number of clipped tokens per training step. Loglinear clips the least, indicating less token waste.}
\label{fig:clipped_tokens}
\end{figure}

Finally, Figure~\ref{fig:clipped_tokens} compares the number of clipped tokens across all three methods. The recomputing and sync methods clip significantly more tokens than the loglinear method (A-3PO), indicating they trigger trust region constraints more frequently. The loglinear method shows least clipping behavior, suggesting smoother, more stable policy updates that naturally stay within trust region bounds, indicating higher sample-efficiency.

\subsubsection{Benchmark Evaluation}

Following common practice for math reasoning tasks, all benchmark evaluations in this section use greedy decoding (temperature $=0$) via lighteval\footnote{\url{https://github.com/huggingface/lighteval}}, making evaluation deterministic.

Table~\ref{tab:results_summary} summarizes the final evaluation rewards and total training times across all methods and setups. In Setup 1, all three methods achieve comparable final evaluation rewards (0.791--0.797) on GSM8K test dataset, with loglinear completing training in 1.53 hours compared to 1.82 hours for recompute (1.2$\times$ speedup) and 2.36 hours for sync (1.5$\times$ speedup). In Setup 2, the asynchronous methods (recompute and loglinear) significantly outperform the sync baseline in both final reward on DAPO-Math-17k test dataset (0.623--0.627 vs 0.443) and training time. Notably, loglinear method completes training in 14.54 hours compared to 16.10 hours for recompute (1.1$\times$ speedup) and 26.15 hours for sync (1.8$\times$ speedup), while maintaining comparable performance to recompute. These results demonstrate that our approximation method consistently delivers efficiency benefits across model scales while maintaining or improving task performance.

\begin{table}[htb]
\caption{Final evaluation reward and training time. Setup 1: Qwen2.5-1.5B-Instruct on GSM8K. Setup 2: Qwen3-8B on DAPO-Math-17k.}
\label{tab:results_summary}
\centering
\begin{tabular}{ll|c|c}
\hline
\textbf{Setup} & \textbf{Method} & \textbf{Final Eval Reward} & \textbf{Training Time (hours)} \\
\hline
Setup 1 & Sync GRPO     & 0.793 & 2.36 \\
        & Recompute & 0.797 & 1.82 \\
        & Loglinear  (A-3PO) & 0.791 & \textbf{1.53} \\
\hline
Setup 2 & Sync GRPO     & 0.443 & 26.15 \\
        & Recompute & 0.627 & 16.10 \\
        & Loglinear (A-3PO) & 0.623 & \textbf{14.54} \\
\hline
\end{tabular}
\end{table}

\begin{table}[htb]
    \caption{Benchmark evaluation for Setup 2: Qwen3-8B on DAPO-Math-17k.}
    \label{tab:benchmark_evaluation}
    \centering
    \begin{tabular}{llccc}
        \hline
        Method & AIME24 pass@1 & MATH500 pass@1 & Average \\
        \hline
        Sync GRPO & $40.00 \pm 9.10\%$ & $46.80 \pm 2.23\%$ & $43.40\%$ \\
        Recompute & $66.67 \pm 8.75\%$ & $62.80\pm 2.16\%$ & $64.74\%$ \\
        Loglinear  (A-3PO) & $\bm{66.67 \pm 8.75\%}$ & $\bm{66.60 \pm 2.11\%}$ & $\bm{66.64\%}$ \\
        \hline
    \end{tabular}
\end{table}

Furthermore, we evaluate the trained model from Setup 2 on two challenging mathematical reasoning benchmarks: AIME2024\footnote{\url{https://huggingface.co/datasets/math-ai/aime24}} and MATH500\footnote{\url{https://huggingface.co/datasets/math-ai/math500}}. Table~\ref{tab:benchmark_evaluation} presents the pass@1 accuracy  for each method on these benchmarks. The results demonstrate that our proposed A-3PO method (loglinear) achieves the best performance over the ``recompute'' and synchronous baseline.

\section{Conclusion}

Decoupled policy optimization algorithms have proven effective for handling data staleness in asynchronous RL systems, but their reliance on explicit proximal policy computation creates a computational bottleneck that limits their practical benefits. In this work, we addressed this limitation through a simple yet principled observation: the proximal policy serves merely as a trust region anchor between the behavior and target policies, and thus does not require expensive neural network evaluation. Instead, it just needs to lie somewhere in between.

Our staleness-aware approximation method implements this insight by interpolating between behavior and target policies in log-probability space, with fresher data weighted more heavily. We evaluated our approach across two experimental setups spanning different model scales: Qwen2.5-1.5B-Instruct on GSM8K and Qwen3-8B on DAPO-Math-17k. Comparing against both the explicit recompute baseline and synchronous GRPO baseline, our results demonstrate substantial practical benefits: up to 1.8$\times$ speedup in training time while maintaining comparable task performance across all three methods. More importantly, our approximation exhibits improved training stability compared to explicit computation, with more controlled importance weights and fewer clipped tokens. Notably, at larger model scales (Qwen3-8B), the recompute method exhibits very high importance weights indicating instability, while our approximation maintains stable behavior, suggesting that simpler can indeed be better and more reliable.

Beyond computational efficiency, this work highlights a broader principle: when designing RL algorithms for large-scale systems, we should question which components truly require expensive computation and which can be approximated from first principles. Our method applies to any decoupled policy optimization approach, not just PPO, making asynchronous RL more practical for training large language models and other computationally demanding domains.

\appendix
\section{Theoretical Analysis}
\phantomsection
\label{app:theoretical}

\begin{theorem}[Sandwich Property and Contractive Stability of Staleness-Aware Proximal Policy Approximation]
Let $\pi_{\mathrm{behav}}(\cdot | s)$ be the behavior policy, $\pi_\theta(\cdot | s)$ be the target policy, and define the staleness-aware proximal policy
\begin{equation}
\log \pi_{\mathrm{prox}}(a | s)
=
\alpha \log \pi_{\mathrm{behav}}(a | s)
+
(1-\alpha)\log \pi_\theta(a | s),
\quad
\alpha =
\begin{cases}
1, & d = 0,\\
\frac{1}{d}, & d \ge 1,
\end{cases}
\end{equation}
where $d$ denotes policy staleness.

Define the importance ratios
\begin{equation}
w(a | s) = \frac{\pi_\theta(a | s)}{\pi_{\mathrm{behav}}(a | s)},
\qquad
r(a | s) = \frac{\pi_\theta(a | s)}{\pi_{\mathrm{prox}}(a | s)}.
\end{equation}

Assume $w(a | s)$ has finite second moment under $\pi_{\mathrm{behav}}(\cdot | s)$.
Then the following properties hold for all states $s$ and actions $a$:
\begin{enumerate}

    \item \textbf{Trust-region sandwich property}
    \begin{equation}
    \min\{\pi_{\mathrm{behav}}(a | s), \pi_\theta(a | s)\}
    \le
    \pi_{\mathrm{prox}}(a | s)
    \le
    \max\{\pi_{\mathrm{behav}}(a | s), \pi_\theta(a | s)\}.
    \end{equation}

    \item \textbf{Contractive stability}
    \begin{equation}
    r(a | s) = w(a | s)^{\alpha}
    \end{equation}
    In particular, $\lim_{d \to \infty} r(a | s) = \lim_{\alpha \to 0} r(a | s) = 1$.

For variance, it vanishes as the staleness increases:
    \begin{equation}
    \lim_{d \to \infty}
    \mathrm{Var}_{a \sim \pi_{\mathrm{behav}}}
    \bigl[r(a | s)\bigr]
    =
    0.
    \end{equation}

\end{enumerate}
\end{theorem}

\begin{proof}[Sandwich property]
For any $x,y>0$ and $\alpha \in [0,1]$,
\begin{equation}
\min(x,y)
\le
x^{\alpha}y^{1-\alpha}
\le
\max(x,y),
\end{equation}
substituting $x=\pi_{\mathrm{behav}}(a | s)$ and $y=\pi_{\mathrm{theta}}(a | s)$ establishes the sandwich property of $\pi_{\mathrm{prox}}$.

Q.E.D.
\end{proof}

\begin{proof}[Contractive stability]
Exponentiating the definition of $\pi_{\mathrm{prox}}$ yields
\begin{equation}
\pi_{\mathrm{prox}}(a | s)
=
\pi_{\mathrm{behav}}(a | s)^{\alpha}
\pi_\theta(a | s)^{1-\alpha}.
\end{equation}

Therefore,
\begin{equation}
r(a | s)
=
\frac{\pi_\theta(a | s)}{\pi_{\mathrm{prox}}(a | s)}
=
\left(
\frac{\pi_\theta(a | s)}{\pi_{\mathrm{behav}}(a | s)}
\right)^{\alpha}
=
w(a | s)^{\alpha},
\end{equation}
which proves the contractive form of the update and $\lim_{d \to \infty} r(a | s)=1$.

To prove the vanishing variance, first, observe the pointwise convergence. For any fixed $w > 0$, as $\alpha \to 0$, we have $w^\alpha \to 1$. Consequently,
\begin{equation}
    \lim_{\alpha \to 0} (w^\alpha - 1)^2 = 0.
\end{equation}

To interchange the limit and the expectation, we apply the Dominated Convergence Theorem (DCT). We must find an integrable random variable $Y$ such that $(w^\alpha - 1)^2 \le Y$ for all $\alpha \in (0, 1]$. We propose the dominating function $Y = (w - 1)^2$. We verify this bound in two cases:
\begin{itemize}
    \item \textbf{Case 1: $w \ge 1$.} Since $\alpha \in (0, 1]$, the function $f(x) = x^\alpha$ is concave and bounded by the identity for $x \ge 1$. Thus, $1 \le w^\alpha \le w$. Subtracting 1 yields $0 \le w^\alpha - 1 \le w - 1$. Squaring both sides, we obtain $(w^\alpha - 1)^2 \le (w - 1)^2$.
    \item \textbf{Case 2: $0 \le w < 1$.} Since $\alpha \in (0, 1]$, the root brings the value closer to 1, i.e., $w \le w^\alpha \le 1$. Thus, the distance to 1 satisfies $|w^\alpha - 1| = 1 - w^\alpha \le 1 - w = |w - 1|$. Squaring both sides yields $(w^\alpha - 1)^2 \le (w - 1)^2$.
\end{itemize}

Therefore, for all $w \ge 0$, we have $(w^\alpha - 1)^2 \le (w - 1)^2$.

Next, we check the integrability of the dominating function. Expanding the term, we have $\mathbb{E}[(w-1)^2] = \mathbb{E}[w^2] - 2\mathbb{E}[w] + 1$. By assumption, the second moment $\mathbb{E}[w^2]$ is finite, which implies $\mathbb{E}[w]$ is also finite. Thus, $\mathbb{E}[(w-1)^2] < \infty$.

Since the sequence converges pointwise to 0 and is dominated by an integrable variable, the DCT implies:
\begin{equation}
    \lim_{\alpha \to 0} \mathbb{E}[(w^\alpha - 1)^2] = \mathbb{E}\left[ \lim_{\alpha \to 0} (w^\alpha - 1)^2 \right] = 0.
\end{equation}
This implies the variance vanishes as $\alpha \to 0$.

Q.E.D.
\end{proof}

\subsubsection*{Acknowledgments}
We are grateful to Yong Zhang, Ge Shi, Zhenan Fan, Vick Wang, Shusheng Xu, and Wei Fu for their assistance in bridging our research with the AReaL community.

\bibliography{mybibliography}
\bibliographystyle{rlj}

\end{document}